# An Improved Grey Wolf Optimization Algorithm for Heart Disease Prediction

Sihan Niu [a,*], Yifan Zhou [b], Zhikai Li [a], Shuyao Huang [c], and Yujun Zhou [d]

[a] Department of Computer Science, Johns Hopkins University, Baltimore MD, USA
[b] Department of Electrical and Computer Engineering, Johns Hopkins University, Baltimore MD, USA
[c] School of Computer Engineering and Science, Shanghai University, Shanghai, China
[d] School of Information Sciences, University of Illinois Urbana-Champaign, Champaign IL, USA
{sniu7, yzhou222, zli264}@jhu.edu, erre1225@shu.edu.cn, yujun3@illinois.edu

**Abstract.** This paper presents a unique solution to challenges in medical image processing by incorporating an adaptive curve grey wolf optimization (ACGWO) algorithm into neural network backpropagation. Neural networks show potential in medical data but suffer from issues like overfitting and lack of interpretability due to imbalanced and scarce data. Traditional Gray Wolf Optimization (GWO) also has its drawbacks, such as a lack of population diversity and premature convergence. This paper addresses these problems by introducing an adaptive algorithm, enhancing the standard GWO with a sigmoid function. This algorithm was extensively compared to four leading algorithms using six well-known test functions, outperforming them effectively. Moreover, by utilizing the ACGWO, we increase the robustness and generalization of the neural network, resulting in more interpretable predictions. Applied to the publicly accessible Cleveland Heart Disease dataset, our technique surpasses ten other methods, achieving 86.8% accuracy, indicating its potential for efficient heart disease prediction in the clinical setting.

## 1    Introduction and Related Work

The use of backpropagation neural networks has recently become a critical strategy to combat heart disease, a leading global health challenge responsible for immense distress and mortality[1]. Various conditions like coronary artery disease, heart attacks, arrhythmias, and heart failure fall under the umbrella of heart disease, affecting populations across the globe. While classical methods for identifying and forecasting heart disease are grounded in manual estimations using determinants such as age, gender, genetic history, blood pressure, cholesterol metrics, and daily habits[2,3], the surge in accessible medical data has paved the way for heightened precision in heart disease prognosis using machine learning (ML).

Machine learning models stand out with their innate ability to sift through expansive medical datasets, reveal concealed correlations, and distinguish intricate associations between multiple risk factors and instances of heart disease[4,5]. By identifying individuals at a heightened risk earlier, one can initiate preventive actions that might mitigate the severity of heart disease, safeguarding individuals, their families, and the broader society. Moreover, the dynamic learning nature of ML models lets them evolve with fresh data, refining their forecasting accuracy with time[6]. This ongoing learning cycle bolsters their proficiency in discerning nuanced patterns and multifaceted links that play a role in heart disease onset and evolution. Armed with these insights, medical professionals can personalize therapeutic strategies, ensuring optimized treatment routes and minimizing redundant tests, thus curbing overall healthcare expenditures.

Several ML methodologies, like logistic regression, support vector machine, KNN, decision tree, random forest, extra trees classifier, gradient boosting, Naive Bayes, and XGBoost, have demonstrated immense promise in forecasting and averting cardiovascular ailments. Timely detection and intervention of heart disease can substantially alleviate its global toll on both individuals and medical infrastructure. The emphasis on predicting heart disease facilitates timely interventions and fosters a proactive preventive approach, enriching patient-centric decisions[7,8,9]. In this landscape of predicting heart disease, numerous groundbreaking algorithms have been put to the test to amplify precision. A prime example is the hybrid algorithm, MLP-PSO, introduced by Ali Al Bataineh and Sarah Manacek[10]. Their trailblazing study merges the strengths of MLP-PSO, offering an innovative way to fine-tune neural network training.

In our study, we introduce an advanced neural network framework for precise heart disease prediction, underpinned by an improved Grey Wolf Optimization algorithm. Drawing from the predatory instincts of grey wolves, the ACGWO methodology fine-tunes the neural network's coefficients and intercepts. Incorporating ACGWO within the backpropagation mechanism, our composite model adeptly navigates the solution domain, boosting the neural network's convergence rate and precision. The consistent refinements in the ACGWO algorithm facilitate effective overall optimization, detecting intricate data patterns associated with heart disease.

## 2    Method

### 2.1    GWO

The Grey Wolf Optimizer (GWO) is a bio-inspired optimization method, introduced by Mirjalili et al.[11]. Drawing from the hierarchical social dynamics and predatory patterns of grey wolves, the GWO aims to thoroughly navigate and pinpoint optimal outcomes within a given problem domain. The procedure initiates by dispersing a group of grey wolves randomly across the domain, with each wolf signifying a possible solution. The methodology then progressively refines the wolves' positions through three main phases: hunting, encircling, and preying. During the hunting phase, wolves modify their placements relative to the alpha, beta, and delta wolf positions[12], maintaining a balance between exploration and exploitation to ensure a holistic navigation of the potential solutions. As the optimization advances, wolf positions undergo continuous adjustments, and their fitness is gauged against the optimization problem's objective function. By methodically refining wolf positions and leveraging the finest located solutions, the algorithm strives to identify the absolute best outcome. While GWO has showcased its efficacy in multiple optimization scenarios, it's pivotal to fine-tune parameters and make context-specific adjustments to maximize its efficiency for distinct challenges.

## 2.2 Cauchy Distribution Using in GWO

The Cauchy distribution, named after the French mathematician Augustin-Louis Cauchy, is a continuous probability distribution that often arises in various fields of statistics and probability theory. It is also referred to as the Cauchy-Lorentz distribution, as it is closely related to the Lorentzian function in physics. Based on the previously mentioned grey wolf algorithm and Cauchy distribution, we have listed the following Eq 1-4:

$$f(x; x_0, \gamma) = \frac{1}{\pi\gamma} \cdot \frac{\gamma^2}{\gamma^2 + (x - x_0)^2} \tag{1}$$

$$\text{fi\_alpha} = -\frac{a2}{\pi} \cdot \frac{1}{a2^2 + \left(\frac{\text{Alpha\_score}}{f\_avg} - b2\right)^2} \cdot c2 + d2 \tag{2}$$

$$\text{fi\_delta} = -\frac{a2}{\pi} \cdot \frac{1}{a2^2 + \left(\frac{\text{Delta\_score}}{f\_avg} - b2\right)^2} \cdot c2 + d2 \tag{3}$$

$$\text{fi\_beta} = -\frac{a2}{\pi} \cdot \frac{1}{a2^2 + \left(\frac{\text{Beta\_score}}{f\_avg} - b2\right)^2} \cdot c2 + d2 \tag{4}$$

A distinguishing feature of the Cauchy distribution is its pronounced tails[13]. This translates to a comparatively elevated likelihood of witnessing extreme outcomes than in other distributions that have defined moments. As a result, the Cauchy distribution is a go-to for representing data characterized by outliers or data exhibiting heavy-tailed tendencies. It's essential to recognize that the Cauchy distribution lacks certain fundamental properties, including the Central Limit Theorem. Owing to its absence of finite moments, conventional statistical methods centered on mean and variance estimations might be ill-suited for data adhering to the Cauchy paradigm. In spite of its unique characteristics, the Cauchy distribution finds applications across multiple fields such as physics, economics, and engineering. Whether utilized as a standard for hypothesis evaluations or integrated within complex models, its significance is palpable. However, while working with Cauchy-distributed data, one should proceed with caution and consider methods specifically designed for handling heavy-tailed distributions.

## 2.3 S-Shaped Curve

$$f(x) = \frac{1}{1 + e^{-x}} \tag{5}$$

$$w = w_{\text{start}} - (w_{\text{start}} - w_{\text{end}}) \frac{1}{1 + e^{(a-b\cdot t)}} \qquad (6)$$

$$ww = \frac{a}{\pi} \cdot \frac{1}{a^2 + \left(\frac{\text{iter}}{\text{Max\_iteration}} - b\right)^2} \cdot c + d \qquad (7)$$

The S-Shaped curve, commonly known as the sigmoid curve, is frequently observed in logistic regression models. It finds wide applications in various fields such as population growth, plant growth, and virus spread.

The sigmoid and inverse sigmoid functions are shown in Eq 5 and Eq 6, respectively. From Eq 5, it can be observed that as $x$ increases, $y$ also increases. Initially, $y$ shows rapid growth, followed by slower growth in the later stages. In this equation, an improved inertia weight linearly decreasing strategy for particle swarm optimization is proposed.

The modified inertia weight formula is given as shown in Eq 6. The S-Shaped curve, commonly known as the sigmoid curve, is frequently observed in logistic regression models. It finds wide applications in various fields such as population growth, plant growth, and virus spread. The Sigmoid function is expressed as where $a$ and $b$ are constants[14]. By setting $x = t$, the expression for the inverse sigmoid function is obtained. In Eq 7, $t$ denotes the ongoing iteration, while $Max\_iteration$ signifies the maximum iteration count for a particle. A reference study carried out a comparative evaluation of four inertia weight functions: linear decrease, downward parabolic curve, upward parabolic curve, and decreasing exponential curve. The outcomes suggest that using the upward parabolic or decreasing exponential curve to model inertia weight can amplify the optimization capability of the algorithm in distinct measures.

Drawing from these findings, multiple strategies have emerged in recent times aimed at bolstering the efficacy of particle swarm optimization. Examples include the Gaussian function decreasing inertia weight approach, the logarithmic decrease method, and the Cauchy distribution-based inertia weight decline technique[15-17]. While such modifications to inertia weight can indeed elevate the particle swarm optimization's performance, they occasionally prompt premature convergence. This happens due to an overly rapid decrease in inertia weight in the primary search phases, leading to a diminished diversity amongst the particle population.

In a bid to counteract these challenges, our investigation taps into the attributes of the inverse S-shaped function for inertia weight adjustments. As shown in Eq 7, this function gradually diminishes across its span but exhibits slower shifts during the commencement and conclusion phases. Such a modification ensures that the inertia weight retains a relatively robust value in the initial phases, fortifying the algorithm's global search potential. Conversely, in the latter stages, by sustaining a reduced inertia weight for an extended duration, the algorithm's local search prowess is augmented.

## 2.4     Training BP Using GWO

Merging the GWO algorithm with the training regimen of neural networks yields several benefits. Primarily, the efficiency of GWO in navigating the search space enables the neural network to bypass local optima, leading to superior solutions. Moreover, the GWO's adeptness at striking a balance between exploration and exploitation aids the neural network in adapting well to novel data, thus minimizing overfitting. Furthermore, with the GWO's inclusion, the training mechanism enhances its resilience, lessening the odds of stagnation in less-than-ideal solutions. To sum up, integrating Grey Wolf Optimization with neural network training rooted in backpropagation presents a formidable method to bolster the network's efficacy in navigating intricate optimization and pattern discernment challenges. By capitalizing on the optimization prowess of the GWO algorithm, neural networks can witness heightened precision and adaptability. This amalgamated strategy paves the way for addressing intricate challenges across computer vision, pattern identification, and beyond. Future studies can delve into refining and diversifying this integration, propelling the discipline forward and addressing intricate real-world dilemmas.

## 2.5 ACGWO Algorithm for Neural Network Backpropagation

As shown in Table 1, the ACGWO algorithm addresses these issues by introducing an adaptive strategy during the update of grey wolf positions. The Adaptive Curve Grey Wolf Optimizer algorithm employs search agents (wolves) that iterate over the search space to find the optimal solution. With each iteration, the positions of the wolves are adaptively updated using calculated parameters such as $A$, $C$, and $D$, which dictate their movement. The algorithm concludes by returning the position of the Alpha wolf, representing the best-found solution.

**Table 1.** Pseudocode for the Adaptive Grey Wolf Optimizer

| Algorithm | Adaptive Curve Grey Wolf Optimizer |
| --- | --- |
| **Require:** | Size of search space, number of search agents, maximum iterations |
| 1: | Initialize the position array Positions of the search agents (wolves) randomly within the problem space. |
| 2: | **for** $iter$ = 1 to $maximum\ iterations$ do |
| 3: | Calculate the fitness of each search agent, update the position and fitness of $Alpha$, $Beta$, and $Delta$. |
| 4: | **for** each $Wolf$ in all $Wolves$ do |
| 5: | Compute new $A$, $C$, $D$, $wa$, and $ww$ values |
| 6: | $A = 2 \times wa \times r1 - wa$ |
| 7: | $C = 2 \times r2$ |
| 8: | $D = C \times Wolf\_position - Current\_position$ |
| 9: | $wa = 2 - iter \times \frac{2}{Max\_iteration}$ |
| 10: | $ww = \frac{aa}{\pi} \times \frac{1}{aa^2 + \left(\frac{iter}{Max\ iteration} - b\right)^2} \times c + d$ |
| 11: | Calculate the new position of the search agent |
| 12: | $X\_new = ww \times Wolf\_position - A \times D$ |
| 13: | Trim the new position to the border of the search space if it exceeds the search space |

| 14: | Update the position of the search agent with the new position |
|-----|---------------------------------------------------------------|
| 15: | Update the position of the search agent with the new position |
| 16: | **end for** |
| 17: | Update the fitness of Alpha with the current best solution |
| 18: | **end for** |
| 19: | **Output:** Return the position of Alpha as the best solution found. |

## 3 Experimental Results

### 3.1 The Comparison and Analysis of 9 Algorithms

The goal of this experiment is to compare the fitness of the ACGWO algorithm with other well-known algorithms such as WDO, GSA, SCA, and PSO. We also compared it to the traditional GWO algorithm, the standalone Adaptive GWO algorithm, and the S-Shaped function GWO algorithm. The 6 basic test functions are depicted in the following Table 2:

**Table 2.** Six benchmark test functions.

| Function No. | Function Name | Range | Optimal Value |
|:------------:|:-------------:|:-----:|:-------------:|
| f1 | Sphere | $[-100, 100]$ | 0 |
| f2 | Schwefel P2.21 | $[-100, 100]$ | 0 |
| f3 | Schwefel P2.22 | $[-10, 10]$ | 0 |
| f4 | Rosenbrock | $[-10, 10]$ | 0 |
| f5 | Quadric Noise | $[-1.28, 1.28]$ | 0 |
| f6 | Schafers | $[-100, 100]$ | 0 |

In the experiment, we set the same population size as n+40, using search dimensions of 30, 100, and 500. Learning factors were set as $c_1=c_2=2$. All other settings are consistent with the original literature, including wmax=0.9 and wmin=0.4. The results are presented in the following Table 3:

**Table 3.** The CGWO and AGWO are enhanced versions of the GWO algorithm. The CGWO incorporates an S-shaped function, introducing a curve to the algorithm to expedite the convergence rate. The AGWO is designed based on insights from the Cauchy inequality, enabling each wolf in the algorithm to adaptively converge.

| Function | Algorithm | dim=30 | | dim=100 | | dim=500 | |
|:--------:|:---------:|:-------:|:-------:|:-------:|:-------:|:-------:|:-------:|
| | | MEAN | STD | MEAN | STD | MEAN | STD |
| f1 | PSO | 7.49E+02 | 5.37E+01 | 2.87E+03 | 1.78E+02 | 1.57E+04 | 3.83E+02 |
| | SCA | 1.52E+01 | 1.39E+01 | 2.65E+02 | 1.46E+02 | 3.18E+03 | 1.39E+03 |
| | GSA | 3.75E+01 | 2.34E+01 | 3.05E+02 | 5.20E+01 | 1.82E+03 | 1.03E+02 |
| | WDO | 2.59E-07 | 3.83E-07 | 1.17E-05 | 1.38E-05 | 1.36E-04 | 2.07E-04 |
| | GWO | 7.84E-06 | 5.45E-06 | 6.67E-02 | 3.93E-02 | 4.45E+01 | 7.47E+00 |
| | CGWO(Ours) | 1.40E-92 | 3.16E-92 | 5.57E-88 | 4.28E-88 | 1.89E-84 | 1.80E-84 |
| | AGWO(Ours) | 4.94E-131 | 1.08E-130 | 3.82E-127 | 1.25E-126 | 1.44E-124 | 1.10E-124 |
| | ACGWO(Ours) | **6.01E-219** | **0.00E+00** | **5.63E-214** | **0.00E+00** | **2.21E-209** | **0.00E+00** |
| f2 | PSO | 7.08E+12 | 1.54E+13 | 1.03E+55 | 4.54E+55 | 3.56E+278 | Inf |
| | SCA | 3.10E+00 | 2.79E+00 | 3.26E+01 | 1.52E+01 | 1.47E+02 | 7.45E+01 |
| | GSA | 4.65E+01 | 9.90E+00 | 1.94E+02 | 1.88E+01 | 1.77E+271 | inf |
| | WDO | 2.81E-03 | 3.51E-03 | 1.60E-02 | 1.30E-02 | 1.60E-01 | 1.23E-01 |
| | GWO | 3.55E-03 | 1.72E-03 | 8.65E-01 | 1.59E-01 | 6.74E+01 | 6.18E+00 |
| | CGWO(Ours) | 5.51E-47 | 3.59E-47 | 4.34E-44 | 3.24E-44 | 6.53E-42 | 1.73E-42 |
| | AGWO(Ours) | 1.14E-65 | 6.64E-66 | 1.18E-63 | 3.91E-64 | 9.77E-62 | 4.82E-62 |
| | ACGWO(Ours) | **2.68E-109** | **5.79E-109** | **5.64E-107** | **4.67E-107** | **1.43E-104** | **1.49E-104** |
| f3 | PSO | 1.04E+03 | 1.53E+02 | 8.83E+03 | 1.06E+03 | 1.70E+05 | 5.01E+04 |
| | SCA | 3.11E+02 | 1.78E+02 | 4.30E+03 | 1.67E+03 | 1.09E+05 | 1.95E+04 |
| | GSA | 2.51E+02 | 9.93E+01 | 1.44E+03 | 4.51E+02 | 2.80E+03 | 1.19E+04 |
| | WDO | 1.47E-05 | 2.15E-05 | 2.20E-04 | 3.84E-04 | 6.76E-03 | 1.03E-02 |
| | GWO | 1.58E+00 | 1.80E+00 | 3.95E+02 | 1.57E+02 | 1.33E+04 | 1.68E+03 |
| | CGWO(Ours) | 5.26E-83 | 2.32E-82 | 1.28E-78 | 4.87E-78 | 1.54E-75 | 4.80E-75 |
| | AGWO(Ours) | 5.21E-130 | 6.37E-130 | 2.00E-127 | 2.75E-127 | 4.71E-125 | 6.18E-125 |
| | ACGWO(Ours) | **5.83E-218** | **0.00E+00** | **3.35E-215** | **0.00E+00** | **5.05E-211** | **0.00E+00** |
| f4 | PSO | 2.22E+01 | 2.46E+00 | 5.38E+01 | 4.56E+00 | 7.74E+01 | 3.85E+00 |
| | SCA | 6.47E+01 | 9.63E+00 | 9.53E+01 | 1.85E+00 | 9.94E+01 | 2.45E-01 |
| | GSA | 3.04E+01 | 6.17E+00 | 4.07E+01 | 4.11E+00 | 6.96E+01 | 1.98E+00 |
| | WDO | 2.91E-03 | 2.07E-03 | 6.45E-03 | 5.40E-03 | 6.28E-03 | 5.30E-03 |
| | GWO | 7.97E-01 | 5.80E-01 | 3.65E+01 | 9.16E+00 | 8.10E+01 | 4.37E+00 |
| | CGWO(Ours) | 2.66E-44 | 2.26E-44 | 7.04E-40 | 1.49E-39 | 3.36E-33 | 4.89E-33 |
| | AGWO(Ours) | 4.15E-64 | 2.76E-64 | 7.18E-63 | 3.87E-63 | 1.62E-61 | 1.39E-61 |
| | ACGWO(Ours) | **1.19E-107** | **1.28E-107** | **8.32E-106** | **1.30E-105** | **1.72E-104** | **1.31E-104** |
| F5 | PSO | 4.70E+02 | 1.82E+01 | 1.69E+03 | 5.43E+01 | 8.88E+03 | 1.17E+02 |
| | SCA | 1.06E+02 | 4.87E+01 | 4.41E+02 | 1.35E+02 | 1.61E+03 | 5.21E+02 |
| | GSA | 2.60E+02 | 3.83E+01 | 9.46E+02 | 8.19E+01 | 4.76E+03 | 1.51E+02 |
| | WDO | 1.22E+02 | 2.94E+01 | 5.19E+02 | 2.69E+02 | 8.75E+02 | 1.80E+03 |
| | GWO | 2.67E+01 | 1.03E+01 | 1.88E+02 | 8.05E+01 | 1.71E+03 | 2.47E+02 |
| | CGWO(Ours) | **0.00E+00** | **0.00E+00** | **0.00E+00** | **0.00E+00** | **0.00E+00** | **0.00E+00** |
|  | AGWO(Ours) | **0.00E+00** | **0.00E+00** | **0.00E+00** | **0.00E+00** | **0.00E+00** | **0.00E+00** |
|  | ACGWO(Ours) | **0.00E+00** | **0.00E+00** | **0.00E+00** | **0.00E+00** | **0.00E+00** | **0.00E+00** |
| F6 | PSO | 1.57E+00 | 9.66E-02 | 1.26E+01 | 1.92E+00 | 6.23E+02 | 3.34E+01 |
|  | SCA | 1.29E+01 | 9.41E+00 | 3.39E+02 | 1.80E+02 | 3.07E+03 | 8.47E+02 |
|  | GSA | 3.52E+02 | 5.50E+01 | 1.85E+03 | 1.25E+02 | 1.22E+04 | 3.01E+02 |
|  | WDO | 4.96E-02 | 1.54E-01 | 3.51E-04 | 4.80E-04 | 1.01E-03 | 2.60E-03 |
|  | GWO | 3.51E-02 | 4.37E-02 | 1.01E+00 | 7.51E-02 | 4.18E+01 | 1.16E+01 |
|  | CGWO(Ours) | **0.00E+00** | **0.00E+00** | **0.00E+00** | **0.00E+00** | **0.00E+00** | **0.00E+00** |
|  | AGWO(Ours) | **0.00E+00** | **0.00E+00** | **0.00E+00** | **0.00E+00** | **0.00E+00** | **0.00E+00** |
|  | ACGWO(Ours) | **0.00E+00** | **0.00E+00** | **0.00E+00** | **0.00E+00** | **0.00E+00** | **0.00E+00** |

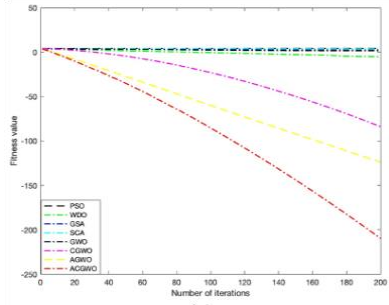

(a)

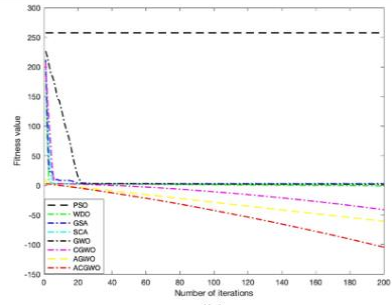

(b)

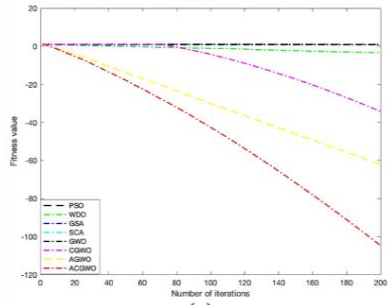

(c)

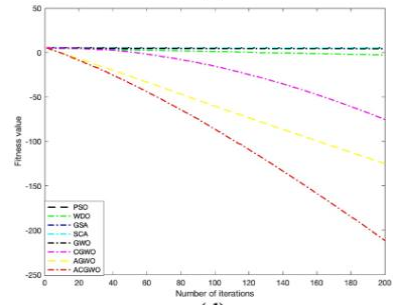

(d)

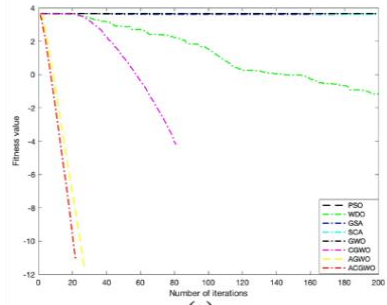

(e)

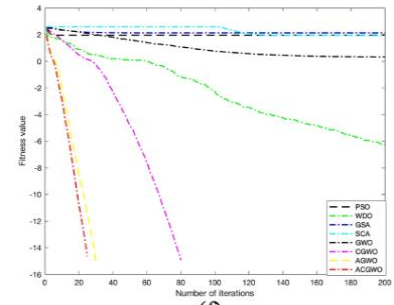

(f)

**Fig. 1.** The loss fitness value curve. (a) to (f) represent the loss fitness values of f1 to f6, respectively, as the number of iterations changes at a dimension of 500.

Fig 1 illustrates that for both unimodal functions (f1-f4) and multimodal functions (f5-f6), the ACGWO algorithm's standard deviation is zero. This data points to the ACGWO algorithm's stability being superior to that of the other four adjusted algorithms. Concurrently, during the tests on multimodal functions from f4 to f6, our enhanced algorithm remains robust, even as the other five algorithms find themselves trapped in local optima. Another observation from Fig 1 is the relatively consistent evolutionary trajectory of the algorithm despite increasing iteration counts.

In side-by-side tests involving just the AGWO algorithm and the exclusive S-Shaped function GWO approach that we included, it became evident that solely relying on the adaptive curve expedites the convergence rate. However, introducing the S-Shaped curve garners even better outcomes, most noticeably with the f5 function test. Keeping in mind the frequent application of the f5 Rastrigin function in neural network weight optimization, we plan to merge the ACGWO algorithm with neural networks in forthcoming experiments and choose pertinent datasets for exploration.

## 3.2 The Comparison and Analysis of 9 Algorithms

In this section, the ACGWO has been applied to diagnose heart disease. The Cleveland heart disease dataset has been used to evaluate the performance of the ACGWO and other 6 models. Before constructing the models, the dataset undergoes thorough preprocessing and cleaning steps, including addressing missing values, converting categorical variables, and applying feature scaling using the standardization method. By employing these techniques, we aim to enhance the accuracy and effectiveness of the diagnostic models for heart disease. Table 4 below shows the optimal parameter settings for Eq 2-4 and Eq 7, determined after multiple experimental attempts.

**Table 4.** Parameters

| Parameter | GWO | | S-Shape | | | | Cauchy | | | |
|---|---|---|---|---|---|---|---|---|---|---|
| | Swarm Size | Iterations | a | b | c | d | a | b | c | d |
| Value | 100 | 1000 | 1.0 | 0.0 | 2.0 | 1.7 | 1.0 | 0.0 | 2.0 | 2.1 |

In Table 4, swarm size represents the number of individuals in the search space; Iterations determine the number of iterations the algorithm will run.

## 3.3 Data Analysis

Exploratory Data Analysis (EDA) is a critical process of conducting preliminary investigations on data. EDA is a significant step in fine-tuning the given dataset in a different form of analysis to understand the insights of the key characteristics of the dataset's various entities. A visual representation of the correlation matrix of the Cleveland heart disease dataset is shown in Figure 2:

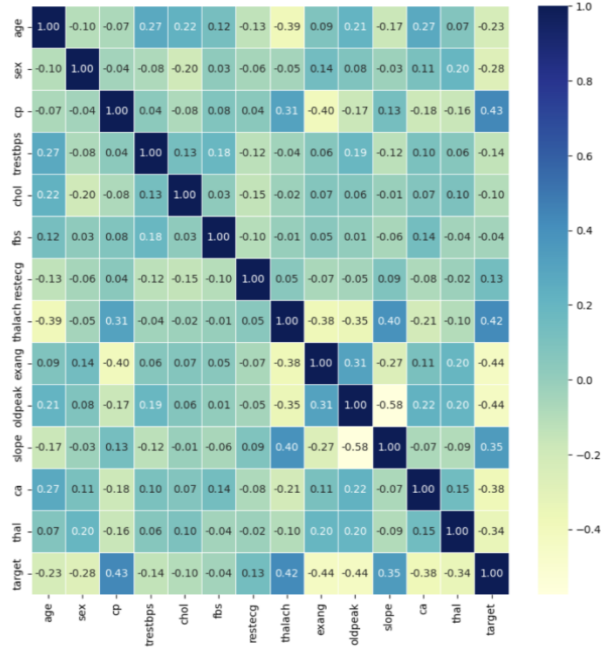

**Fig. 2.** Correlationtrix for heart.csv

Figure 2 displays the pairs with the strongest correlation. Given that pairs with high correlation essentially depict the same variance in the dataset, analyzing these pairs can aid in identifying the more significant attribute within each pair for model development. Intriguingly, no individual feature in the dataset exhibits a strong correlation with the desired outcome. In addition, there are varied correlations among the features in relation to the target value; some show a negative correlation while others demonstrate a positive correlation, as illustrated in Eq 8.

$$r_{xy} = \frac{\sum_{i=1}^{n}(x_i - \bar{x})(y_i - \bar{y})}{\sqrt{\sum_{i=1}^{n}(x_i - \bar{x})^2 \sum_{i=1}^{n}(y_i - \bar{y})^2}} \tag{8}$$

### 3.4 Data Preprocessing

Data preprocessing is a fundamental step in machine learning, ensuring raw data is refined and structured, paving the way for effective model training. Essentially, this technique in data mining converts raw data into a more digestible and interpretable format. In this study, our data preprocessing steps encompass: Managing Null or Missing Values; Processing Categorical Data; Segmenting Data into Training and Testing Sets; and Scaling Features.

### 3.5 Performance Evaluation Metrics

In the evaluation of our proposed diagnostic models for heart disease, we employed rigorous performance metrics to assess their effectiveness. We use the same dataset as Ali Al Bataineh and Sarah Manacek [10], a stratified 70-30 train-test split was employed, where 70% of the data was used for training, and the remaining 30% was held out for testing across all models. The performance metrics used to evaluate the effectiveness of the diagnostic models are as follows:

$$\text{Accuracy} = \frac{TP + TN}{TP + TN + FP + FN} \tag{9}$$

$$\text{Precision} = \frac{TP}{TP + FP} \tag{10}$$

$$\text{Recall} = \frac{TP}{TP + FN} \tag{11}$$

$$\text{F1 Score} = \frac{2 \times \text{Precision} \times \text{Recall}}{\text{Precision} + \text{Recall}} \tag{12}$$

### 3.6    Result and Discussion

The objective of this research was to craft a diagnostic model for heart disease employing machine learning methodologies. Multiple models, with the ACGWO being a notable mention, underwent training, fine-tuning, and assessment processes. Findings from experiments conducted on the Cleveland heart disease dataset can be found in Table 5.

**Table 5.** Performance Comparison of ML Classifiers[10]

| Algorithms | ACC | AUC | PRE | Recall | F1 |
|---|---|---|---|---|---|
| Gradient Boosting | 0.714 | 0.712 | 0.725 | 0.659 | 0.690 |
| Decision Tree | 0.758 | 0.756 | 0.775 | 0.681 | 0.738 |
| Extra Trees | 0.769 | 0.766 | 0.810 | 0.681 | 0.740 |
| XGB | 0.769 | 0.766 | 0.810 | 0.681 | 0.740 |
| KNN | 0.780 | 0.777 | 0.815 | 0.704 | 0.756 |
| Random Forest | 0.791 | 0.787 | 0.857 | 0.681 | 0.759 |
| MLP | 0.802 | 0.799 | 0.861 | 0.704 | 0.775 |
| Logistic Regression | 0.813 | 0.808 | 0.909 | 0.681 | 0.779 |
| SVM | 0.813 | 0.809 | 0.885 | 0.704 | 0.784 |
| GaussianNB | 0.824 | 0.821 | 0.868 | 0.750 | 0.804 |
| MLP-PSO | 0.846 | 0.848 | 0.808 | 0.883 | 0.704 |
| ACGWO-BP(Ours) | **0.868** | **0.863** | **0.854** | **0.880** | **0.870** |

In the above graph. We have found that ACGWO has shown the best results. That might be due to several factorials that lead to this result. We have discussed this in further context.

## 4 Conclusion

We initiated this research by addressing two main challenges faced by the GWO algorithm: its proneness to local optima and its occasional slow convergence for certain problems. Our approach to improving the convergence rate incorporated a Sigmoid function, which led to noteworthy enhancements. An adaptive strategy was adopted to combat the local optima challenge. Drawing inspiration from the foundational principles of GWO, which replicates the dynamics of wolf packs, we integrated a Cauchy distribution, chosen for its relevant physical implications. The refined algorithm, when tested across six functions, showed superior performance compared to the traditional GWO, evidencing a combined benefit. Our enhanced algorithm, termed ACGWO, expanded the capabilities of the original GWO, especially in equipment and detection use cases.

In the context of this research, we introduced an ACGWO-optimized BP neural network tailored for diagnosing cardiovascular diseases. We delved into various potential machine-learning models to craft intelligent cardiovascular diagnostic systems. Using the Cleveland dataset, which comprises 13 distinct features to determine the likelihood of heart disease, we conducted exhaustive experiments. Our findings were illuminating: the proposed ACGWO-BP model surpassed all other tested algorithms, achieving an impressive accuracy of 86.8%. This underscores the potential of the ACGWO framework in aiding medical professionals to make more precise diagnoses and prescribe optimal treatment pathways.

## References

1. Al Bataineh, A.; Jarrah, A. High Performance Implementation of Neural Networks Learning Using Swarm Optimization Algorithms for EEG Classification Based on Brain Wave Data. Int. J. Appl. Metaheuristic Comput. 2022, 13, 1–17. [CrossRef]
2. Tsao, C.W.; Aday, A.W.; Almarzooq, Z.I.; Alonso, A.; Beaton, A.Z.; Bittencourt, M.S.; Boehme, A.K.; Buxton, A.E.; Carson, A.P.; Commodore-Mensah, Y.; et al. Heart Disease and Stroke Statistics—2022 Update: A Report From the American Heart Association. Circulation 2022, 145, e153–e639. [CrossRef] [PubMed]
3. Petersen, K.S.; Kris-Etherton, P.M. Diet Quality Assessment and the Relationship between Diet Quality and Cardiovascular Disease Risk. Nutrients 2021, 13, 4305. [CrossRef]
4. Samieinasab, M.; Torabzadeh, S.A.; Behnam, A.; Aghsami, A.; Jolai, F. Meta-Health Stack: A new approach for breast cancer prediction. Healthc. Anal. 2022, 2, 100010. [CrossRef]
5. Hameed, B.Z.; Prerepa, G.; Patil, V.; Shekhar, P.; Zahid Raza, S.; Karimi, H.; Paul, R.; Naik, N.; Modi, S.; Vigneswaran, G.; et al. Engineering and clinical use of artificial intelligence (AI) with machine learning and data science advancements: Radiology leading the way for future. Ther. Adv. Urol. 2021, 13, 17562872211044880. [CrossRef] [PubMed]
6. Oikonomou, E.K.; Williams, M.C.; Kotanidis, C.P.; Desai, M.Y.; Marwan, M.; Antonopoulos, A.S.; Thomas, K.E.; Thomas, S.; Akoumianakis, I.; Fan, L.M.; et al. A novel machine learning-derived radiotranscriptomic signature of perivascular fat improves cardiac risk prediction using coronary CT angiography. Eur. Heart J. 2019, 40, 3529–3543. [CrossRef] [PubMed]

7. Kumar, Y.; Koul, A.; Singla, R.; Ijaz, M.F. Artificial intelligence in disease diagnosis: A systematic literature review, synthesizing framework and future research agenda. J. Ambient. Intell. Humaniz. Comput. 2022, 1–28. [CrossRef] [PubMed]

8. Al Bataineh, A. A comparative analysis of nonlinear machine learning algorithms for breast cancer detection. Int. J. Mach. Learn. Comput. 2019, 9, 248–254. [CrossRef]

9. Doppala, B.P.; Bhattacharyya, D. A Novel Approach to Predict Cardiovascular Diseases Using Machine Learning. In Machine Intelligence and Soft Computing; Springer: Berlin/Heidelberg, Germany, 2021; pp. 71–80.

10. Ali Al Bataineh, & Sarah Manacek. (2022). MLP-PSO Hybrid Algorithm for Heart Disease Prediction. Journal of Personalized Medicine, 12(8), 1208. https://doi.org/10.3390/jpm12081208

11. Mirjalili, S., Mirjalili, S. M., & Lewis, A. (2014). Grey Wolf Optimizer. Advances in Engineering Software, 69, 46-61. DOI: 10.1016/j.advengsoft.2013.12.007

12. Gupta, S., & Deep, K. (2019). A novel Random Walk Grey Wolf Optimizer. Swarm and Evolutionary Computation, 44, 101-112. DOI: DOI: 10.1016/j.swevo.2018.06.013

13. Ghosh, J., Li, Y. B., & Mitra, R. (2018). On the use of Cauchy prior distributions for Bayesian logistic regression. Bayesian Analysis, 13(2), 359-383.

14. GU, Y., LU, H., XIANG, L., & SHEN, W. (2022). Adaptive Simplified Chicken Swarm Optimization Based on Inverted S-shaped Inertia Weight. Chinese Journal of Electronics, 31(2).

15. RATNAWEERA, A., HALGAMAGE, G., & WATSON, H. C. (2004). Self-organizing hierarchical particles warm optimizer with time-varying acceleration coefficients [J]. IEEE Transactions on Evolutionary Computation, 8(3), 240-255.

16. CLERC, M., & KENNEDY, J. (2002). The particle swarm-explosion, stability, and convergence in a multidimensional complex space [J]. IEEE Transactions on Evolutionary Computation, 6(1), 58-73.

17. Lillicrap, T. P., Santoro, A., ..., & Hinton, G. (2020). Backpropagation and the brain. Nature Reviews Neuroscience, 21(6), 335-346.